\documentclass{article}

\usepackage{PRIMEarxiv}

\usepackage[utf8]{inputenc} 
\usepackage[T1]{fontenc}    
\usepackage{hyperref}       
\usepackage{url}            
\usepackage{booktabs}       
\usepackage{amsfonts}       
\usepackage{nicefrac}       
\usepackage{microtype}      
\usepackage{lipsum}
\usepackage{fancyhdr}       
\usepackage{graphicx}       
\graphicspath{{media/}}     
\usepackage{biblatex} 
\usepackage{amsmath,amssymb}
\usepackage{bm}

\DeclareMathOperator*{\argmax}{arg\,max}

\title{Dealing with Sparse Rewards Using Graph Neural Networks
\thanks{This work was supported in part on Section 2 by the Strategic Project ‘‘Digital Business’’ within the framework of the Strategic Academic Leadership Program ‘‘Priority 2030’’ at the National University of Science and Technology (NUST) MISiS, in part by the Basic Research Program at the National Research University Higher School of Economics (HSE University), and in part by the Computational Resources of HPC Facilities at HSE University.

\textit{\underline{Citation}}: 
\textbf{M. Gerasyov and I. Makarov, "Dealing With Sparse Rewards Using Graph Neural Networks," in IEEE Access, vol. 11, pp. 89180-89187, 2023, doi: 10.1109/ACCESS.2023.3305927.}} 
}

\author{
  Matvey Gerasyov \\
  School of Data Analysis and Artificial Intelligence, HSE University, Moscow, Russia \\
  Laboratory of Algorithms and Technologies for Network Analysis, HSE University, Nizhny Novgorod, Russia \\
  \texttt{msgerasyov@hse.ru} \\
   \And
  Ilya Makarov \\
  Laboratory of Algorithms and Technologies for Network Analysis, HSE University, Nizhny Novgorod, Russia \\
  AI Center, NUST MISiS, Moscow, Russia \\
  Artificial Intelligence Research Institute (AIRI), Moscow, Russia\\
  \texttt{makarov@airi.net} \\
}

\begin{document}
\maketitle

\begin{abstract}
Deep reinforcement learning in partially observable environments is a difficult task in itself and can be further complicated by a sparse reward signal. Most tasks involving navigation in three-dimensional environments provide the agent with minimal information. Typically, the agent receives a visual observation input from the environment and is rewarded once at the end of the episode. A good reward function could substantially improve the convergence of reinforcement learning algorithms for such tasks. The classic approach to increasing the density of the reward signal is to augment it with supplementary rewards. This technique is called reward shaping. In this study, we propose two modifications of one of the recent reward shaping methods based on graph convolutional networks: the first involving advanced aggregation functions, and the second utilizing the attention mechanism. We empirically validate the effectiveness of our solutions for the task of navigation in a 3D environment with sparse rewards. For the solution featuring the attention mechanism, we can also show that the learned attention is concentrated on edges corresponding to important transitions in the 3D environment.

\end{abstract}


\section{Introduction}
Reinforcement learning is a machine learning paradigm where an artificial agent learns the optimal behavior through interactions with a dynamic environment. Goals and purposes are explained to the agent via a scalar reward signal it receives after each interaction.  Throughout the training process, the agent infers the behavior that maximizes cumulative reward, also called the return. To succeed in this task, the agent needs to explore the environment to understand which states and actions yield high rewards. On the other hand, the agent also has to exploit the rewards it has already received to adapt its behavior. This problem is known as the exploration and exploitation trade-off. 

Deep reinforcement learning (DRL) algorithms use neural networks to process large or continuous state spaces. The deep reinforcement learning approach has proven worthy in many dynamic tasks, such as machine translation \cite{https://doi.org/10.48550/arxiv.1808.08866, Satija2016SimultaneousMT, 10.4218/etrij.2020-0358}, robotics \cite{mahmood2018benchmarking, 8832393, surmann2020deep}, playing videogames \cite{hessel2017rainbow, wydmuch2018vizdoom, openai2019dota, 020makarov2017learning,045makarov2019deep,047makarov2019deep,050makarov2019deep,065makarov2021deep}, and performing navigation in complex environments \cite{https://doi.org/10.48550/arxiv.1612.03801, gym_miniworld, 005makarov2015imitation,009makarov2016smoothing,010makarov2016modelling,014makarov2016first}. In addition to these domains, deep reinforcement learning has demonstrated significant potential for solving real-world control problems, such as predictive aircraft maintenance \cite{LEE2023108908} and traffic signal control \cite{9241006, li2021deep}.

Navigating in three-dimensional environments can present a challenging problem for agents due to the sparsity of rewards. This problem arises when a scant number of states in the state space return a meaningful reward signal. A typical situation is when the agent must find a specific item or place in the environment and receives a positive reward only after reaching the destination. From the RL training procedure formulation, it naturally follows that one wants to reward the agent as often as possible. Hence, sparse rewards are detrimental to learning efficiency.

Throughout recent years several papers have addressed the sparse reward problem. Some notable approaches include reward shaping, Curiosity-Driven Methods \cite{pathak2017curiositydriven, 8961529, NEURIPS2021_1e8ca836}, Curriculum Learning \cite{narvekar2020curriculum, 9874056, https://doi.org/10.48550/arxiv.2006.09641}, Adaptive Skill Acquisition \cite{10.1007/978-3-030-61616-8_31, 10.1007/978-3-030-86380-7_53}, and learning with Auxiliary Tasks \cite{jaderberg2017reinforcement, riedmiller2018learning, NEURIPS2019_0e900ad8}. This study focuses on the potential-based reward shaping technique, as it is the most straightforward and intuitive way to deal with the sparse reward problem. This method is very flexible since it can be combined with most general-purpose RL algorithms. 

This paper proposes a novel modification to a recently developed reward shaping technique based on the message-passing mechanism of graph convolutional networks \cite{klissarov2020reward}. Over recent years, graph neural networks have become increasingly popular and have found their application across various domains, including reinforcement learning \cite{Wei_2019, gammelli2021graph}. As a result, numerous graph neural network architectures have emerged, offering different benefits \cite{Wu_2021}. We show how selecting the appropriate architecture can notably increase the quality of the learned shaping function. For this purpose, we conduct several experiments using environments with sparse rewards from MiniWorld \cite{gym_miniworld}.

\section{Background and Motivation}
Table \ref{tab:rl_params} summarizes the parameters, variables, and functions used throughout this paper.

\begin{table}[h] 
\centering 
\caption{Summary of notations used in the paper} 
\begin{tabular}{|l|l|} 
\hline 
Notation & Description \\
\hline 
$\mathcal{S}$ & State space \\ 
$\mathcal{A}$ & Action space \\
$\mathcal{R}$ & Reward function \\
$s, s_t \in \mathcal{S}$ & Current state \\
$s' \in \mathcal{S}$ & Next state \\
$a, a_t \in \mathcal{A}$ & Current action \\
$r$ & Immediate reward \\ 
$\pi(a|s)$ & Policy\\ 
$\pi^*$ & Optimal policy\\ 
$\gamma$ & Discount factor \\
$G_t$ & Discounted return at timestep $t$ \\
$V(s)$ & State-value function\\
$A(s,a)$ & Advantage function\\
$J_{actor}$ & Objective of the actor \\
$r_t(\theta)$ & Probability ratio between the policies\\
$clip(r_t(\theta), 1 - \epsilon, 1 + \epsilon)$ & Clip $r_t(\theta)$ between $1 - \epsilon$ and $1 + \epsilon$\\
$\mathcal{R}'(s, a, s')$ & Shaped reward function \\
$F(s, a, s')$ & Shaping function \\
$\Phi$ & Scalar potential function defined on states \\
$h$ & Node embeddings \\
$\mathcal{N}(i)$ & Set of neighbors of node $i$ \\
$\mathbb{S}$ & Set of base case states \\
$A$ & Adjacency matrix of the graph \\
$X$ & Matrix of node features \\
$\mathcal{L}_0$ & Cross entropy component of the loss\\
$\mathcal{L}_{prop}$ & Message-passing component of the loss\\
\hline 
\end{tabular} 
\label{tab:rl_params} 
\end{table}

\subsection{Deep Reinforcement Learning Overview}
Markov decision process (MDP) is a standard model of agent-environment interaction. An MDP is a tuple $(\mathcal{S}, \mathcal{A}, \mathcal{P}, \mathcal{R})$, where $\mathcal{S}$ is a finite state space and $\mathcal{A}$ is a finite action space. $\mathcal{P}$ denotes a state transition function, giving the transition probability $p(s_{t+1} \mid s_t, a_t)$. Finally, $\mathcal{R}$ is a scalar reward function. A fundamental property of MDP is that the conditional probability distribution given by $\mathcal{P}$ depends only on the current state and does not depend on the history of the process. Discounted return is the sum of all rewards starting from state $s_t$ multiplied by a discount factor $\gamma \in [0, 1)$:
\begin{equation}
    G_t = \sum_{k=0}^{\infty} \gamma^k r_{t + k + 1},
\end{equation}
where $r_{t + k + 1}$ is the reward recieved at timestep $t + k + 1$.

In the partially observable MDP setting, the states are not entirely observable by the agent, introducing additional challenges to reinforcement learning algorithms.

At each step, an agent takes a decision according to a policy $\pi(a|s)$. The main goal of reinforcement learning is to find an optimal policy $\pi^*$ that maximizes the expected discounted return:
\begin{equation}
    \pi^* = \argmax_{\pi} \mathbb{E}_{\pi} [G_0]
\end{equation}
Value function $V_\pi(s)$ is the expected discounted return conditional on the state of the environment: 
\begin{equation}
    V_{\pi}(s) = \sum_a \pi(a \mid s) \sum_{r, s'} p(r, s' \mid s, a) [r + \gamma V_\pi(s')]
\end{equation}
where $s'$ is the next state after state $s$, $r$ and $a$ are the reward and action at the current step respectively. $v_{\pi}(s)$ represents the value of following policy $\pi$ from state $s$.

Policy-based DRL methods approximate the agent's policy with a neural network. One of the most famous policy-based methods is called advantage actor-critic \cite{mnih2016asynchronous}. The underlying neural network has two heads called actor and critic, respectively. The actor learns an optimal policy, and the critic learns the value function, which represents the quality of a given state. Thus, the information provided by the critic helps train the actor. The update rule for the actor reflects this fact:
\begin{equation}
\label{eqn:jactor}
    \nabla_{\theta} J_{actor} = \frac{1}{N} \sum_{t=0}^{N} 
    \nabla_{\theta} \log \pi_{\theta} (a_t \mid s_t) A(s_t, a_t)
\end{equation}
where $\theta$ denotes the parameters of the network, $N$ is the number of steps in the trajectory, and $A$ is the advantage of action $a$ in state $s$. The advantage function represents the quality of a chosen action compared to the expected baseline, given by the value function. Thus, following the update rule given by (\ref{eqn:jactor}), we aim to increase the probability of choosing beneficial actions with positive advantage values. One can estimate the advantage from a part of a trajectory $(s, a, r, s')$ of an agent as follows:
\begin{equation}
    \hat{A}(s, a) = r + \gamma V(s') - V(s)
\end{equation}
Training the critic can be formulated as a regression problem with the following loss function:
\begin{equation}
    L_{critic} = \frac{1}{N} \sum_{i=0}^{N} \sum_{s, a}
    (V_{\theta}(s) - [r + \gamma V(s')])^2
\end{equation}

One problem with this approach is that the value of the learning rate does not guarantee the degree of policy change. Small changes in the network parameters can lead to abrupt changes in the quality of a policy. The Proximal Policy Optimization (PPO) algorithm \cite{schulman2017proximal} addresses this issue by minimizing the following objective:
\begin{equation}
    \hat{\mathbb{E}}_t 
    [min(r_t(\theta)\hat{A}_t, clip(r_t(\theta), 1 - \epsilon, 1 + \epsilon)\hat{A}_t)]
\end{equation}
where $r_t(\theta)$ is the probability ratio $r_t(\theta) = \frac{\pi_{\theta}(a_t \mid s_t)}{\pi_{\theta_{old}}(a_t \mid s_t)}$, and clip denotes clipping $r_t$, which removes the benefit of moving $r_t$ outside of the interval $[1 - \epsilon, 1 + \epsilon]$.

\subsection{Reward shaping}
The reward shaping framework aims to solve the sparse reward problem by augmenting the original reward function with a shaping function $F(s, a, s')$:
\begin{equation}
    \mathcal{R}'(s, a, s') = \mathcal{R}(s, a, s') + F(s, a, s')
\end{equation}
Shaping functions can be hand-crafted based on expert knowledge of the problem (e.g., Euclidean distance to the goal) or inferred during the training procedure \cite{klissarov2020reward,article}. The necessary and sufficient condition for preserving the set of optimal policies of an MDP is for the shaping function to take the following form \cite{Ng99policyinvariance}:
\begin{equation}
    F(s, a, s') = \gamma \Phi(s') - \Phi(s)
\end{equation}
where $\Phi$ is the scalar potential function defined on states. Designing a good potential-based shaping by hand can be a challenging task for some problems. Moreover, hand-crafted reward shapings often lack performance compared to automatically learned ones \cite{klissarov2020reward}.

\subsection{Graph neural networks}

A graph is a data structure representing a set of objects and relations between them. A graph $G$ can be defined as a collection of nodes and edges connecting them $G = \{V, E\}$. Each node and edge of the graph can store additional information in the form of feature vectors. Graphs are an essential tool for modeling heterogeneously structured data, such as MDPs in reinforcement learning problems. Graph neural networks (GNNs) are deep learning models that allow for inference on graphs by leveraging local graph structure and node-level features. Graph convolutional networks (GCNs) are a special kind of GNN that implement a message-passing mechanism through the aggregation of neighboring nodes' features. In this section, we discuss commonly-used graph convolutional models. 

The original Graph Convolutional Network \cite{kipf2017semisupervised} performs aggregation of neighboring nodes' features normalized by node degrees. The output of one convolutional layer of such a network can is defined as follows:
\begin{equation}
    h_i^{(l+1)} = \sigma(b^{(l)} + \sum_{j \in \mathcal{N}(i)} \frac{1}{c_{ji}}h_j^{(l)}W^{(l)}),
\end{equation}
where $\mathcal{N}(i)$ is the set of neighbors of node $i$, $c_{ji} = \sqrt{|\mathcal{N}(j)|} \sqrt{|\mathcal{N}(i)|}$, $W$ is a learnable weight matrix, $l$ is the number of layer, and $\sigma$ is a non-linear activation function.

Graph Attention Network (GAT) \cite{gat2018graph} adds the attention mechanism to GCN. GAT convolution aggregates node features of neighbors proportional to attention scores $a_{i, j}$:
\begin{gather}
 h_i^{(l+1)} = \sigma(\sum_{j \in \mathcal{N}(i)} a_{i, j} W^{(l)} h_j^{(l)}); \\
 a_{i, j}^{(l)} = softmax_{i}(e_{i, j}^{(l)});\\
 e_{i, j}^{(l)} = LeakyReLU(\vec{a}^{(l)^T}[W h_i^{(l)} || W h_j^{(l)}]),
\end{gather}
where $||$ denotes concatenation and $\vec{a}^{(l)}$ is a learnable weight vector.

Finally, the GraphSAGE model \cite{hamilton2018inductive} allows using different aggregators, such as mean, pooling, and LSTM \cite{lstm}:
\begin{gather}
    h^{(l+1)}_{\mathcal{N}(i)} = aggregate \big(\{h_j^l, \forall j \in \mathcal{N}(i) \} \big); \\
    h_i^{(l+1)} = \sigma\big(W \cdot concat(h^l_i, h^{(l+1)}_{\mathcal{N}(i)}) \big),
\end{gather}
where $aggregate$ is one of the aggregators from the list mentioned above and $concat$ stays for the concatenation of node embeddings.

The key distinction among the outlined models lies in how they propagate messages between nodes to update their embeddings. Choosing an appropriate aggregating procedure can strongly affect the performance of graph convolutional networks. Next, we discuss learning an optimal potential-based reward shaping using graph convolutional networks.

\subsection{Reward Propagation Using Graph Convolutional Networks}
In \cite{klissarov2020reward}, the authors propose applying GCNs to a graph in which each state is a node and edges represent a possible transition between two states.  Since there is no access to the complete underlying graph, it is approximated through sampled trajectories. The key idea of this approach is to propagate information about rewarding states through the message-passing mechanism implemented by GCNs. The probability distribution $\Phi_{GCN}$ at the output of the GCN is used as a potential function for potential-based reward shaping. To train the GCN, the authors use the following loss function:

\begin{gather}
    \mathcal{L} = \mathcal{L}_0  + \eta \mathcal{L}_{prop}\\
    \mathcal{L}_0 = \sum_{s \in \mathbb{S}} p(O \mid s) \log (\Phi_{GCN}(s)) \\
    \mathcal{L}_{prop} = \sum_{v, w} A_{vw} \|\Phi_{GCN}(X_w) - \Phi_{GCN}(X_v)\|^2
\end{gather}

Here, $\mathbb{S}$ is the set of base case states, which consists of the first and last states of a trajectory and the states with non-zero rewards. $A$ is the adjacency matrix, and $X$ is the matrix of node features. $\mathcal{L}_0$ is the cross entropy loss between the labels of states from $\mathbb{S}$ and predictions of the GCN model. $\mathcal{L}_{prop}$ combines the neighboring messages through the graph Laplacian. $\eta$ is the hyperparameter controlling the contribution of $\mathcal{L}_{prop}$ component to the whole loss. Lesser values of $\eta$ lead to a more simple and biased model. Following the original paper, $\eta$ is set to be equal to $10$ in all the experiments.

\section{Related Work}
\subsection{Applications of deep reinforcement learning}
Deep reinforcement learning has been successfully applied to real-world problems in various domains. For example, in \cite{a15110393}, the authors employed Deep Q-Learning \cite{mnih2013playing} to predict lithium-ion battery capacity based on the permutation entropy of battery voltage sequences. Similarly, deep reinforcement learning was utilized for predictive aircraft maintenance \cite{LEE2023108908}. The authors used a Soft-Actor-Critic \cite{haarnoja2018soft} agent to decide when to schedule an engine replacement based on the estimates of Remaining-Useful-Life. Another domain where reinforcement learning has shown promising results is traffic signal control \cite{9241006}. For instance, in \cite{zheng2019learning}, the authors adopted the distributed framework of Ape-X DQN \cite{horgan2018distributed} to learn a generalizable policy for operating a signalized intersection. In addition, deep reinforcement learning has been effectively applied to mobile robot navigation in indoor environments \cite{surmann2020deep}. The authors trained an A3C \cite{mnih2016asynchronous} agent using only data from a 2D laser scanner and an RGB-D camera. These examples demonstrate the potential of deep reinforcement learning to improve decision-making in complex systems.

\subsection{Graph convolutional networks}

Graph convolutional networks have been a significant development in the field of representation learning on graphs. Since their introduction \cite{kipf2017semisupervised}, multiple modifications have emerged to enhance their performance. One area of research has focused on developing more effective methods of neighborhood aggregation. For instance, GAT \cite{gat2018graph} has added an attention mechanism to GCN, while GraphSAGE \cite{hamilton2018inductive} adapted the aggregation process to incorporate advanced aggregators such as Long Short-Term Memory (LSTM) \cite{lstm}. However, the choice of the optimal architecture heavily depends on the task at hand, and it remains an active research topic \cite{dwivedi2022benchmarking}. Model quality can be affected by various design choices, including the style of message passing, the number of message-passing layers, the dimensionality of embeddings, layer connectivity, and others \cite{you2021design}. In addition, different architectures vary in their expressive power \cite{xu2019powerful}. Therefore, further research is necessary to discover better architectures suitable for emerging tasks, such as applying graph neural networks in reinforcement learning.

\section{Proposed Approach}
Following the reward propagation framework discussed previously, we propose using GAT and GraphSAGE models to propagate information about rewarding states. We use two-layered implementations of these models with 64 hidden units. The first layer of our GAT model has four attention heads and LeakyReLU activation function. As for the GraphSAGE model, we use mean aggregation and ReLU activation in the first layer. We chose the LSTM aggregator for the second layer since it is more appropriate considering the sequential nature of the data in reinforcement learning. We use an actor-critic network with a three-layered CNN encoder to model the policy of all agents. The architecture of the encoder is provided in Table~\ref{tab1}. 
\begin{table}[ht]
\centering
\caption{CNN encoder architecture.}\label{tab1}
\begin{tabular}{|c|c|c|c|c|}
\hline
Layer &  Number of filters & Kernel size & Stride & Activation\\
\hline
1 &  32 & 8 & 4 & ReLU\\
2 &  64 & 4 & 2 & ReLU\\
3 &  32 & 3 & 1 & ReLU\\
\hline
\end{tabular}
\end{table}

The number of attention heads in the GAT model was tuned by comparing the results of multiple experiments. Other design choices highlighted in the previous section are consistent with the prior work \cite{klissarov2020reward} to make a fair comparison. We argue that the proposed models have an inductive bias which can help the models leverage the specific structure of the data in the problem at hand. The GraphSAGE model with the LSTM Aggregator can take advantage of the sequential nature of the data in the agent's trajectory. The GAT model can learn which transitions in the underlying MDP are relevant to the agent's task due to the attention mechanism. In addition, the GAT model provides interpretability as we can directly evaluate the learned attention scores. The training data for the models are samples of the underlying MDP graph represented by linear graphs of the agent's trajectories. This sampling strategy has been demonstrated to be a valid technique for training graph convolutional networks, as it does not result in a significant deterioration of model performance \cite{10.1145/1150402.1150479, klissarov2020reward}.

The forward path in GCNs involves an aggregation of neighboring nodes’ features. A simple approach to implementing the aggregation step is to use a matrix multiplication between the adjacency and feature matrices. However, this method has a high time complexity of $O(N^2F)$, where $N$ represents the number of nodes and $F$ is the dimensionality of node embeddings. It is possible to take advantage of the sparse nature of adjacency matrices in the problem at hand. By utilizing sparse operators, the time complexity of the aggregation can be reduced to linear with respect to the number of nodes \cite{blakely2021time}. This approach offers significant improvements, enabling efficient processing of large graphs. Thus, the time complexity of the forward path in GCNs is linear with respect to the number of nodes in the framework of the considered problem. The proposed models don't involve any expensive matrix operations such as inversion. They only require a constant number of additional matrix multiplications per layer when compared to GCN used in prior work \cite{klissarov2020reward}. Given that all models in our experiments have the same number of layers and hidden units, the proposed approach can be considered comparable in computational time to the baseline architecture.

The PPO algorithm is used to update the policy in all experiments. The node features provided to GNNs come from the output of the CNN encoder of the actor-critic network. Finally, we compare our agents, denoted $\Phi_{GAT}$ and $\Phi_{GraphSAGE}$ respectively, with two baselines: $\Phi_{GCN}$ introduced in the original paper \cite{klissarov2020reward} and basic PPO without any reward shaping.

\section{Experiment Design and Results}

\begin{figure}[ht]
\centering
\includegraphics[width=0.45\columnwidth]{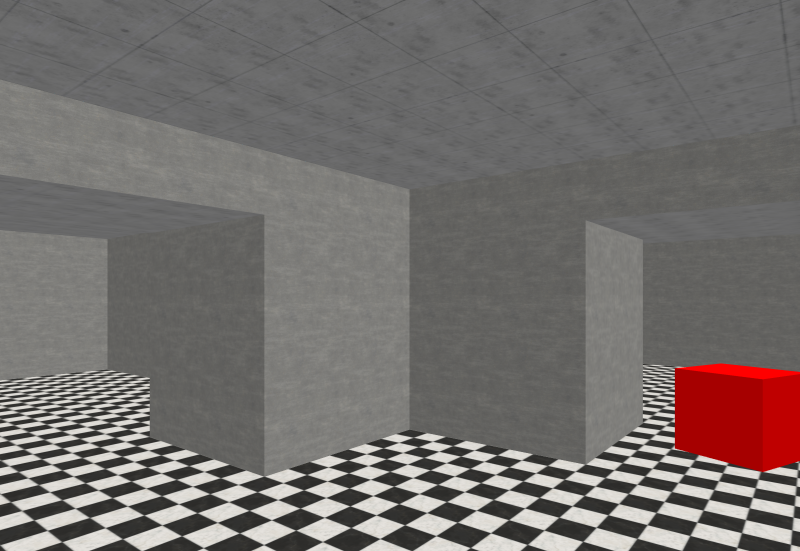}
\hfill
\includegraphics[width=0.45\columnwidth]{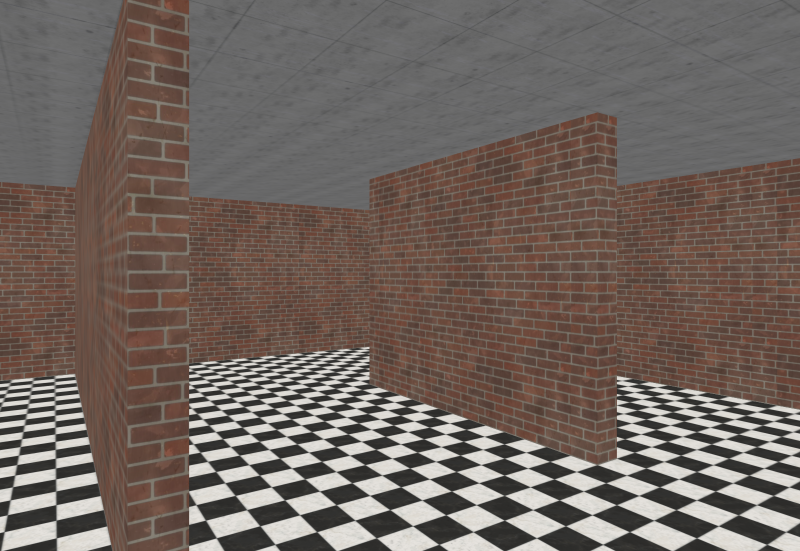}
\caption{Screenshots of the FourRooms environment (left) and Maze environment (right)} 
\label{figscreen}
\end{figure}

We perform a series of experiments in MiniWorld \cite{gym_miniworld} to test our approach. MiniWorld has several challenging three-dimensional POMDP environments.  For our experiments, we select two environments with sparse rewards: FourRooms and Maze. Screenshots of both environments are shown in Figure \ref{figscreen}. In the next two sections, we describe them in detail as well as the training procedure. We state our results in the final section of this chapter.

\subsection{FourRooms}

The player spawns at a random position inside four rooms connected by four openings. In order to get a reward, the player must reach the goal, represented by a red box. Furthermore, the position of the goal is also random for each episode. Also, there is a time limit to perform this task which is 250 steps. The player chooses one of three actions at each step: move forward, turn left, and turn right. The environment provides a positive reward only when the player succeeds. In this case, the reward is scaled down proportional to how long it took the player to reach the goal. The rest of the original rewards are zeros.

We train all neural networks on 16 parallel instances of the environment. We organize training based on the algorithm outlined in \cite{klissarov2020reward}. The node features used by the GNN come from the output of the CNN encoder of the actor-critic network.  Each agent interacts with its environment for 128 steps. During this process, we record hidden states at the output of the CNN encoder and add all transitions $(s_t, s_{t+1})$ to the graph $G_i$, where $i$ is the number of the environment. Then we apply reward shaping using the current potential function $\Phi$ and split the resulting sequence into four mini-batches. Finally, we use these mini-batches and PPO to update the policy. When environment $i$ reaches the end of an episode, we use the recorded hidden states, the set of the base case states $\mathbb{S}_i$, and the graph $G_i$ to update the potential function $\Phi$ at the output of the GNN model. Since an agent does not receive non-zero rewards until the end of an episode, $\mathbb{S}_i$ only consists of the first and last states of a trajectory. We repeat this update procedure until the total number of steps made by agents in all 16 environments exceeds 5 million. 

\subsection{Maze}

The player has to navigate to a goal through a procedurally generated maze. The player and the goal spawn randomly inside this maze, and the action space is the same as in the previous environment. The critical difference that makes this environment much harder than FourRooms is that the map is generated randomly at the beginning of each episode. The maze generation procedure begins at the top-left corner and utilizes a recursive backtracking algorithm to construct the maze. At each step, the algorithm randomly selects a neighboring room that hasn't been visited before and connects it to the current room. If there are no available unexplored rooms, the algorithm backtracks until it finds an unvisited room or returns to the starting position. This process generates a connected acyclic graph, ensuring that every room is reachable from any other room and that there is only one path between any two rooms. Subsequently, walls are placed between neighboring rooms that are found to be disconnected after completing the generation procedure. All walls inside the maze have the same color and texture. After the maze is generated, the goal and the agent are placed in random locations within randomly chosen rooms. The time limit for this environment is 216 steps, and rewards are assigned according to the same rule as in FourRooms. Altogether this makes the Maze environment a very challenging POMDP with sparse rewards.

The MiniWorld developers provide four versions of this environment, each with distinct size and movement characteristics. These versions include MazeS2, which is a small 2 by 2 maze, MazeS3, which is a medium-sized 3 by 3 maze, MazeS3Fast, which has increased turning and moving motion per action, making navigation easier for the agent, and Maze, the largest version with an 8 by 8 size. In this study, we use the MazeS3 version, which has standard movement and turning speeds and consists of 9 interconnected rooms.  The training procedure is the same as described in the previous section, but since this environment is much more challenging, we extend the training duration to 20 million steps.

\subsection{Results}

\begin{figure}[ht]
\centering
\includegraphics[width=0.49\columnwidth]{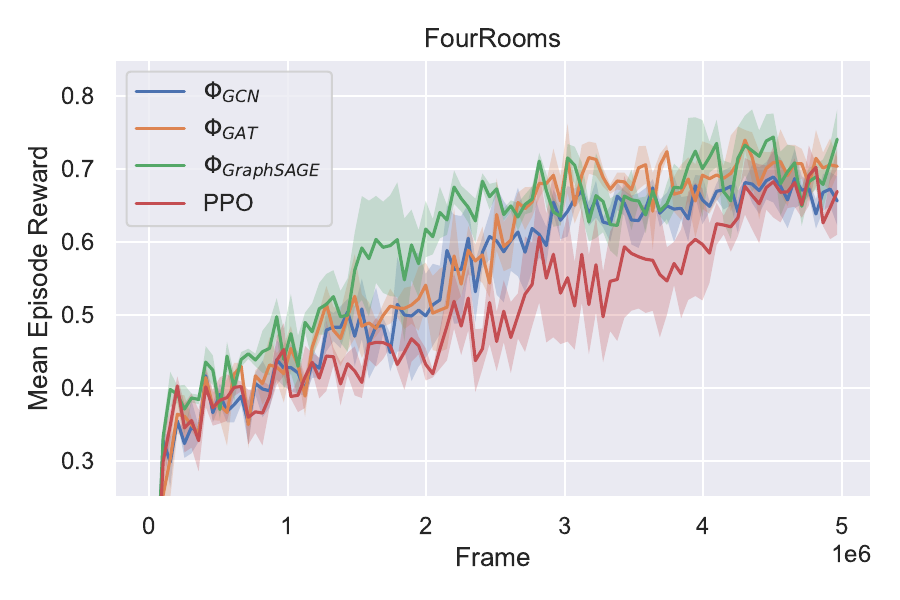}
\hfill
\includegraphics[width=0.49\columnwidth]{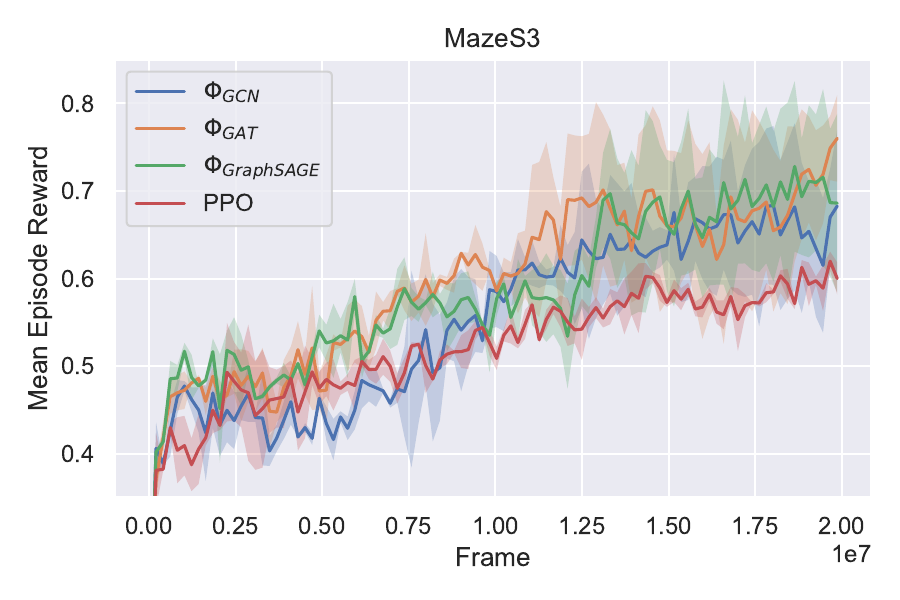}
\caption{Learning curves for the FourRooms environment (left) and MazeS3 environment (right). Results are in the form of $mean(R) \pm std(R)$} \label{fig1}
\end{figure}

The average rewards of the agents during training are shown in Figure ~\ref{fig1}. 

Here, we can immediately see that agents augmented with reward shaping learn faster than baseline PPO. In FourRooms experiments, after 5 million steps, all models converge to a very similar final performance, with $\Phi_{GAT}$ and $\Phi_{GraphSAGE}$ being marginally better. However, in the case of the MazeS3 environment, the difference between the models is more explicit. Also, the $\Phi_{GAT}$ model kept improving even at the end of the training, indicating that the result may be refined.

The final performance of all models is provided in Table~\ref{tab2}. It is worth noting that $\Phi_{GAT}$ has the best mean final performance in both environments. Although, the difference is significant only in the case of a more challenging MazeS3.

\begin{table}[ht]
\centering
\caption{Final performance for both environments. We convert all rewards to $[0, 100]$ scale for better visibility. All results are shown in the form of $mean(R) \pm std(R)$}\label{tab2}
\begin{tabular}{|l|l|l|}
\hline
Model &  FourRooms & MazeS3\\
\hline
$\Phi_{GAT}$ &  $69.93 \pm 5.72$ & $76.56 \pm 5.27$\\
$\Phi_{GraphSAGE}$ &  $69.91 \pm 5.64$ & $68.96 \pm 10.32$\\
$\Phi_{GCN}$ &  $65.28 \pm 5.07$ & $66.96 \pm 9.55$\\
PPO &  $66.79 \pm 6.31$ & $59.75 \pm 4.42$\\
\hline
\end{tabular}
\end{table}

\begin{figure}[b]
\centering
\includegraphics[width=0.5\columnwidth]{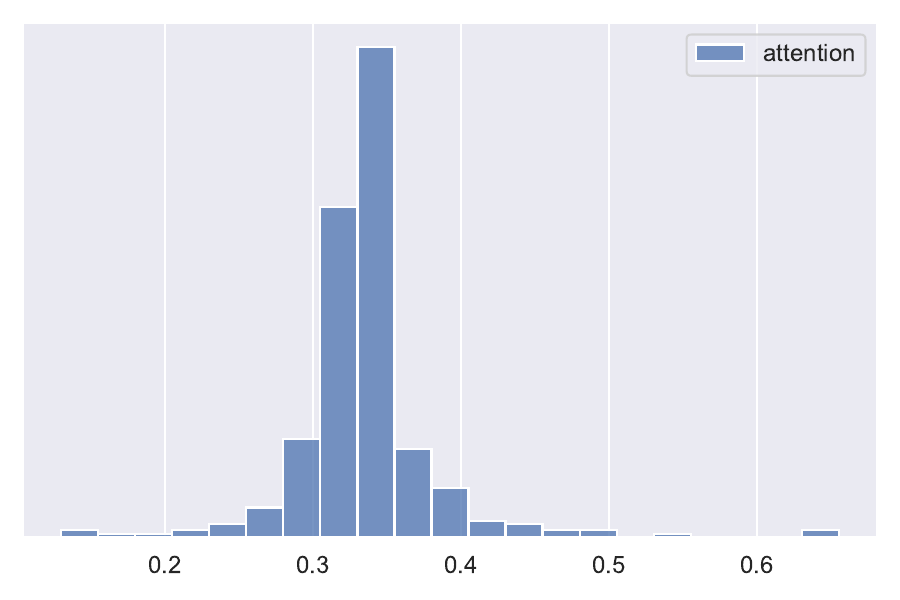}
\hfill
\includegraphics[width=0.45\columnwidth]{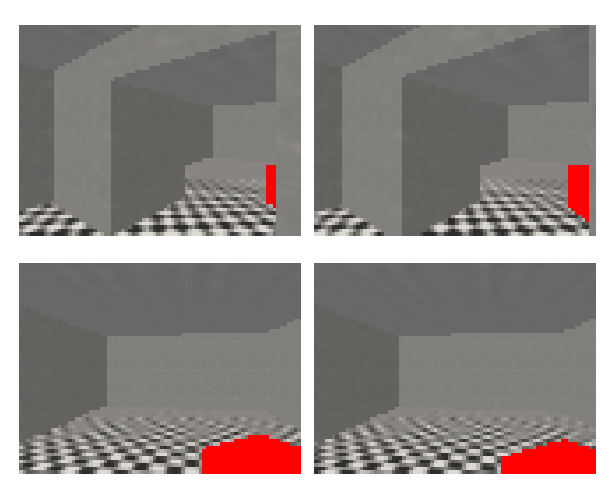}
\caption{Histogram of the learned attention for the FourRooms environment (left) and pairs of screenshots corresponding to the transitions that received high attention (right).}
\label{figatt1}
\end{figure}

Finally, we assess the quality of attention learned by the GAT model. Figure \ref{figatt1} demonstrates the distribution of attention values for the FourRooms environment.

We observe that the learned attention is largely focused on the edges at the beginning of the trajectory and before reaching the goal. Also, high attention is given to the transitions at the point in time when the goal (the red cube) enters the agent's field of view (see Figure \ref{figatt1}). This result, in particular, is unusual since the model did not receive any additional supervision to highlight such edges.

Hence, we can conclude that the GAT model learns to focus its attention on the transitions that are important for the agent's task.

\section{Conclusion}

This study presented two modifications of one of the novel reward shaping techniques. Both our agents demonstrated better convergence speed and final results compared to the baselines. We also showed that the GAT model, which achieved the best final performance for both environments, was also able to learn meaningful attention relevant to the task performed by the agent.

In future work, it may be beneficial to incorporate edge-level or graph-level features. This would provide the graph neural network responsible for learning the potential function with additional information about the environment. Moreover, it may be valuable to explore a more complex design of a transition graph to better capture the structure of the underlying Markov decision process.

\printbibliography 
\end{document}